# Revolutionizing TCAD Simulations with Universal Device Encoding and Graph Attention Networks

Guangxi Fan, Leilai Shao, and Kain Lu Low*

School of Mechanical Engineering, Shanghai Jiao Tong University

800 Dongchuan Road, Shanghai, China, 200240

*Email: kainlulow@sjtu.edu.cn

**Abstract:**

An innovative methodology that leverages artificial intelligence (AI) and graph representation for semiconductor device encoding in TCAD device simulation is proposed. A graph-based universal encoding scheme is presented that not only considers material-level and device-level embeddings but also introduces a novel spatial relationship embedding inspired by interpolation operations typically used in finite element meshing. Universal physical laws from device simulations are leveraged for comprehensive data-driven modeling, which encompasses surrogate Poisson emulation and current-voltage (IV) prediction based on the drift-diffusion model. Both are achieved using a novel graph attention network, referred to as RelGAT. Comprehensive technical details based on the device simulator Sentaurus TCAD are presented, empowering researchers to adopt the proposed AI-driven Electronic Design Automation (EDA) solution at the device level.

## I. Introduction:

As Moore's Law continues to drive the evolution of semiconductor technology, the field faces mounting challenges due to the ever-increasing complexity of designs [1], [2]. The ongoing pursuit of smaller, more efficient devices and advanced fabrication processes has led to heightened requirements for Electronic Design Automation (EDA) tools, necessitating the incorporation of not only new physical models but also accelerated computational methodologies [3], [4]. The concept of AI for EDA has gained substantial traction and found widespread applications in the semiconductor technology field to meet these challenges [5]–[9]. However, device-level EDA, particularly in the context of device design and simulation, remains a relatively underexplored territory.

Leveraging AI techniques for EDA at the device level presents a host of challenges, primarily due to the intrinsic complexity and diversity of device designs, materials, and process variations [10]–[12]. It calls for an effective, universally applicable, and scalable encoding method that captures the nuances of these varying design elements. However, existing EDA tools, particularly commercial device simulation solutions like Sentaurus TCAD, often have complex data file formats, encompassing meshing setup, boundary conditions, material parameters, and physical models. All these make data extraction a non-trivial task. In contrast, in-house software developed by large corporations, although potent, limits accessibility for academic researchers, thereby stifling broader progress in the field [13]–[15].

Among existing encoding methods, three primary types are prevalent. The first type is based on physical features, such as gate length, doping concentration [16]–[18], etc. However, this approach lacks universality due to the significant disparities among different devices. The second type of encoding employs an image-based approach, well-suited for convolutional neural networks (CNN) [13], [14], [19]. However, due to its reliance on uniform grid points, this method struggles with the non-structured grid characteristic of the simulated device, a feature introduced by finite element methods. Consequently, the transformation from non-structured grids to uniform grids often introduces significant interpolation errors. The third type is a graph-based approach that shows promise in encoding the devices with non-structured grids [15], facilitating a seamless representation of non-structured grids. Nevertheless, there remain opportunities for further enhancement. On the one hand, it still requires feature representation, akin to other methods, to effectively encapsulate the unique characteristics of the device. On the other hand, the current feature engineering in the graph-based approach utilizes one-hot vector formatting to capture the type of materials. However, the continuous discovery of new materials and the corresponding expansion of the material library demand lengthening the one-hot vector to accommodate these additions. This poses potential challenges when incorporating new materials into device designs, as it may require frequent updates to the encoding scheme.

In light of these challenges and opportunities, we present innovative formulations designed to transform device-level modeling and computation through a machine learning approach, built on the foundation of a graph-based encoding scheme. Fig. 1 provides a clear overview of these novel contributions, showcasing our advancements in the field of AI-driven EDA at the device level.

Firstly, we propose a graph-based universal encoding approach that provides a more versatile device encoding method. Our approach is designed to be applicable to a wide range of devices, facilitating seamless transferability and scalability across different device types.

Secondly, we have developed an integrative Application Programming Interface (API) that efficiently extracts complex data structures from proprietary commercial software. This API automates the process of establishing the corresponding graph-based encoding output, overcoming the complexities associated with extracting and transforming these data structures. The integrative API has been thoroughly verified through the seamless integration of data structures from Sentaurus TCAD with PyTorch, a widely-used machine learning platform. The integrative API bridges the gap between proprietary commercial software and machine learning framework, facilitating comprehensive analysis and compatibility.

Thirdly, we innovatively proposed a novel approach that utilizes relative position information as an edge feature in graph attention network model (RelGAT) [20]. This strategy, grounded in physical device structure, dynamically assigns importance to nodes based on the relative position information. Drawing from universal physical rules of device modeling, the innovative applications of our methodologies are further demonstrated in two case studies that involve a node-regression-based Poisson emulator and a graph-regression-based drain current emulator using the drift-diffusion equation.

Lastly, our framework emphasizes open accessibility and inclusivity distinguishes it by not relying on proprietary software. This approach fosters a more inclusive research environment, allowing all researchers interested in utilizing AI in the device domain to access and contribute to the advancement of this field.

## II. Approach

### a. Graph-Based Unified Device Encoding

The key foundation of our methodology lies in the graph-based unified device encoding. This approach revolutionizes the representation of devices by leveraging the inherent power of graph structures to capture the complex relationships and dependencies within the devices. The graph-based unified device encoding in Fig. 1 comprises three main components: material-level embedding, device-level embedding, and spatial relationship embedding.

Our methodology introduces material-level embedding, which allows for generalization across diverse devices. It combines a one-hot vector indicating the type of material (metal, insulator, or semiconductor) with a parameter vector that encompasses material properties and physics models. By utilizing material-level embedding, we can capture the fundamental characteristics of different materials and their interactions within the device structure. This encoding approach provides a flexible and scalable representation that accommodates various device types, enabling the application of commonly used physics models such as high field saturation and SRH recombination. The incorporation of physics models further enhances the adaptability of our approach and enables accurate simulations and predictions for different device designs.

In the device embedding process, we incorporate various essential aspects to capture the specific characteristics of individual devices comprehensively. Firstly, we introduce the coordinates of each node within the device, providing spatial information that contributes to the overall representation. Secondly, we consider the bias applied to the entire device, which plays a crucial role in device behavior. Thirdly, we utilize a one-hot vector to indicate the region to which each node belongs, including the gate, drain, source, and body regions. Moreover, the contact properties are also embedded into a vector including contact resistance, contact Schottky barrier, and contact work function. Additionally, we account for the doping concentration in the device by representing the n-type and p-type dopants. Lastly, the operational temperature of the device is also included. By integrating all these data elements, we construct the device embedding, which effectively captures the unique properties and operational characteristics of individual devices.

In our approach, we not only incorporate material-level and device-level embedding but also introduce an innovative technique that simulates interpolation operations commonly used in finite element meshing. This technique embeds the relative positions between nodes as edge features in the graph representation. By including these relative position features, we capture the spatial relationships between different elements within the device structure, akin to the interpolation operations employed in finite element meshing. This integration of spatial relationship embedding, inspired by finite element methods, enhances the encoding process by considering the spatial arrangements and connectivity patterns within the device. It provides valuable context and refines the representation, leading to a more comprehensive understanding of device behavior and enabling more accurate modeling, simulation, and optimization.

Furthermore, our encoding approach incorporates physics-informed AI elements. Device simulation involves solving a self-consistent system of nonlinear equations, requiring the consideration of various self-consistent features in device simulation, such as electrostatic potential and charge density. Therefore, our encoding approach is highly flexible, accommodating the utilization of input features for end-to-end modeling or incorporation of self-consistent features to address specific challenges encountered in device simulations. This flexibility enables the development of data-driven models that effectively capture the complexities of device behavior and accurately represent the intricacies of self-consistent simulations.

In summary, our graph-based unified device encoding methodology, complemented by physics-informed AI elements, offers a versatile and comprehensive approach for representing devices. By combining material-level embedding, device-level embedding, spatial relationship embedding, and physics-informed AL elements within a graph structure, we can capture the intricate relationships and dependencies within the devices. This encoding methodology paves the way for AI-driven advancements in semiconductor device design, enabling more accurate simulations, optimizations, and innovations in the field.

**b. Integrative API bridging Device Simulator and Machine Learning Platform**

This section presents a methodology for creating an API bridge from a device simulator to a machine learning framework, highlighting the independence of our API from proprietary software. We employ a commercial device simulator software, namely Sentaurus TCAD, along with the open-source machine learning platform PyTorch, to develop an integrative API, as shown in Fig. 2.

Sentaurus device simulation provides users with built-in files, such as .tdr and .plt files, containing essential information about the device. The .tdr file includes the device's grid and coordinates information, input characteristics of each node (e.g., doping and temperature), and self-consistent characteristics (e.g., electrostatic potential and charge density). The .plt file contains the device's bias and current results. As the .tdr is a proprietary format, we utilize Sentaurus data explorer to convert the .tdr files into .grd files containing grid information and .tif and .dat files containing characteristics for each node. Additionally, we extract material-related .par files from the Sentaurus material database, which provides material types, material parameters, and physical model parameters. Finally, through automation using Python, we integrate and encode this information into a format compatible with PyTorch, establishing a bridge from TCAD to PyTorch.

To address the data-driven modeling of electrostatic potential computation based on Poisson's equation (equation 1), we incorporate the charge density as an additional feature. Similarly, for the data-driven modeling of current computation based on drift-diffusion and integration over the drain region (equation 2), we include the electrostatic potential and charge density as additional self-consistent features. By considering these additional features in our approach, we aim to enhance the accuracy and reliability of our data-driven models for electrostatic potential and current computations in semiconductor devices. In addition, our approach aims to simulate the universal physical laws governing the device behavior. By incorporating the electrostatic potential, charge density, and other relevant features, we strive to capture the underlying physics that govern the device's operation. This allows us to develop data-driven models that can accurately represent the behavior of a wide range of semiconductor devices, contributing to the understanding and advancement of device-level simulations in the field.

$$\varepsilon \nabla^2 \varphi = -q[p(r) - n(r) + N_D^+ - N_A^-] \tag{1}$$

$$J_n = q\mu_n n E + q D_n \nabla n \tag{2.1}$$

$$J_p = q\mu_p pE - qD_p \nabla p \tag{2.2}$$

$$I_d = \int_{\sigma D} (J_n + J_p) dA \tag{2.3}$$

**c. Graph-based Architecture Design for Diverse Problem Domains**

In a standard graph convolutional network (GCN), each node's feature update is based on the aggregation and combination of its neighbor node features [21]. They employ a fixed mean aggregation to combine neighbor node features, assuming equal importance of neighbor nodes and treating all neighbors equitably. This approach, however, overlooks the importance of neighbor nodes. Due to the mean aggregation of GCN, neighbor node features average each other out, leading to blurred representations and increased similarity. This can reduce the distinction between nodes and limit the model's performance in handling complex tasks. Therefore, in the case of deep GCN networks, over-smoothing phenomena occur [22].

In the context of finite element meshing with unstructured configurations, the proximity of nodes is indicative of their relevance, while distant nodes have relatively diminished influences. Therefore, incorporating relative position as an edge feature, and integrating it into the attention score calculation for neighbor nodes aligns with our understanding [23]. We introduce an improved graph attention network model [20] that incorporates relative position information as an edge feature (RelGAT), a strategy that refines the conventional neighbor-node modeling and employs attention mechanisms to dynamically model the importance of each node's neighbors. Therefore, the importance of node neighbors not only considers the information of the node itself but also takes into account the connecting features between nodes.

We have devised a systematic approach to establish reliable machine learning models across different tasks. In the case of electrostatic potential prediction which is node regression, our initial step was to explore the impact of model complexity on the prediction accuracy. To accomplish this, we employed a combination of shallow GCN and multi-layer perception (MLP) models, collectively known as FatGCN, with varying numbers of parameters to assess their influence on complexity. By analyzing the relationship between the number of model parameters and validation error (as shown in Fig. 3), we were able to identify the optimal model that served as our baseline. We discovered that a model with around 1 million parameters was most suitable.

Based on these findings, we further conducted tests involving both FatGCN models and deeper GCN models to thoroughly evaluate their performance. For comparison, we considered two variations of deep GCN models: regular deep GCN (DeepGCN) and a version with residual connections (ResGCN) that addresses the issue of over-smoothing commonly associated with deeper models [24], [25]. Additionally, we included our RelGAT model in the comparative analysis to ensure a comprehensive assessment. The working modes of these models are depicted in Fig. 4, showcasing the distinct characteristics and mechanisms of the GCN, residual-connected GCN, and GAT with edge features.

Next, we focused on the graph regression task for drain current prediction. For this task, we employed a FatGCN model to test the prediction of drain current. We found that only 15% (0.15 million) of the parameters required for potential prediction were needed to achieve a mean square error (MSE) of 0.43% on the training loss. Given that a FatGCN was sufficient, we only compared the results with those from a RelGAT model.

The architecture of our baseline model is illustrated in Fig. 5. We utilized the Adam optimizer for optimization, and our dataset consisted of 10,000 instances. This dataset was divided into training, validation, and test sets in percentages of 70%, 20%, and 10%, respectively. All hyperparameters were set to the default values and the training processes are shown in Fig. 6.

By following this systematic approach, we were able to assess and compare the performance of different model variations for both node regression and graph regression tasks. The stepwise evaluation allowed us to identify the most suitable modeling approach for each task, considering factors such as model complexity, accuracy, and prevention of overfitting. These findings formed the basis for developing reliable machine learning models for electrostatic potential and drain current predictions in semiconductor devices.

## III. Results and Discussion

In this section, we delve into the comprehensive presentation of our proposed model, RelGAT, building on the baseline architecture discussed earlier. To maintain an equitable comparison, we have also examined the performance of several GCN-based models with identical parameter sizes on the same dataset. In the experimental phase, we employed a dataset encompassing 50,000 samples, with 70% (35,000) earmarked for training, 20% (10,000) allocated for validation, and the remaining 10% (5,000) held for testing. To rigorously evaluate the generalization ability of our model, an additional set of 32,000 unseen samples was randomly selected.

For a more comprehensive understanding and interpretation of the experimental results, we provide a detailed account of the architecture and training specifics of all the models tested. This includes the hierarchical structure of the models, the parameter settings employed, and the optimization algorithms used during the training process. We anticipate that this information will facilitate readers in comprehending our models and experimental results, and serve as a reference point for future related studies.

**a. Node regression-based Poisson emulator**

In terms of the Poisson emulator based on node regression, we have found that a model with a parameter size of 1M is fairly adequate, warranting the exploration of deeper GNNs. Hence, apart from RelGAT, we have made comparisons with three GCN-based network models: FatGCN, DeepGCN, and ResGCN. The network architecture of these models is illustrated in Table 1. Throughout our testing, we utilized the Adam optimizer with a learning rate set to 0.001 and employed Mean Squared Error (MSE) as the loss function. Additionally, we adopted a custom learning rate scheduling scheme, which is a piece-wise linear learning rate decay, as depicted in Fig. 7.

The training processes of the four models are displayed in Fig. 8, from which it can be observed that all models converge to relatively low loss values, with none of them experiencing overfitting issues. Fig. 9 provides an analysis and comparison of the performance of these models on the 50,000-sample dataset as well as an additional unseen dataset of 32,000 samples. It is observed that the RelGAT model surpasses the performance of the GCN-based models.

In addition, according to Table 2, we find that RelGAT exhibits a performance improvement of 72.18% compared to the baseline FatGCN. This is a larger performance boost than DeepGCN's 36.52% and ResGCN's 49.10%, indicating that our RelGAT model has more room for performance enhancement. In the node regression problem for the Poisson emulator, our RelGAT model achieved an R square ($R^2$) score of 99.9977% on 32,000 randomly selected unseen samples. This score indicates that the model can account for almost 100% of the variance, highlighting its exceptional predictive accuracy. This reaffirms the advantage of using graph attention networks that take into account relational information in their attention mechanism, adding to the robustness and efficacy of our proposed RelGAT model. Notably, the electrostatic potential distribution predicted by the RelGAT model, as illustrated in Fig. 10 closely aligns with the results obtained from TCAD simulations, indicating the effectiveness of our model in capturing the essential characteristics of the semiconductor device behavior.

**b. Graph regression-based Drift-Diffusion emulator**

For the prediction of drain current, which is a graph regression problem, our approach reveals that a shallow graph neural network (GNN) is sufficient to achieve a low training loss, as described in the approach section. Based on this insight, we utilized FatGCN and RelGAT to construct two network architectures specifically tailored for graph regression. The network architectures can be referenced in Table 3.

Similar to our previous tests, we employed the Adam optimizer with an initial learning rate of 0.001 for training the models. The loss function used was MSE loss. However, due to the potential instability of MSE when dealing with labels that have a large range, we applied a logarithmic transformation to the current values, which were used as the output of the model. This transformation helps stabilize the learning process and enhances the performance of the models.

To further optimize the training process, a similar custom piece-wise linear learning rate decay schedule employed in the node regression was used to enhance model training convergence. Noticing that graph regression for current prediction tends to encounter overfitting as shown in Fig. 6, we introduced a weight decay of $1\times10^{-4}$ and early stopping to mitigate the overfitting issue. As depicted in Fig. 11, the training curve shows that the performance of RelGAT does not improve beyond 400 epochs. Hence, for RelGAT, we saved the model state at 400 epochs for final testing.

Fig. 12 illustrates the performance of FatGCN and RelGAT on a dataset of 50,000 samples. To verify the generalization ability of the models, we also randomly selected an additional dataset of 32,000 samples for testing. Similar to the previous conclusion, the RelGAT model, which considers relative position as an edge feature, demonstrated superior prediction performance. According to Table 4, RelGAT surpassed the traditional GCN-based models by 12.32% in the graph regression prediction of drain current.

In the evaluation stage for RelGAT, on the randomly drawn additional unseen dataset of 32,000 samples, we also assessed the $R^2$ score of the order of current to be 99.9980, and the $R^2$ score for the absolute magnitude of the current to be 99.4261. These figures indicate a convincingly high level of precision, thus affirming the effectiveness of our unified encoding method. Fig. 13 shows a prediction result of our RelGAT model on drain current, demonstrating a high degree of agreement with TCAD predictions for both drain current in logarithmic and linear scales.

### c. Summary

Overall, our findings reveal that the RelGAT model outperforms the conventional GCN models in both node regression prediction for electrostatic potential and graph regression for current prediction. This indicates a superior level of precision and generalization offered by our RelGAT model when tackling these types of prediction challenges.

For the graph regression problem related to the current prediction, the performance improvement of RelGAT is not as pronounced as that observed in the node regression for Poisson potential prediction. A possible explanation is that we found a small model to be sufficient to achieve a high level of accuracy, with the model size being only 15% of that required for potential prediction. Predicting a single current value might be simpler than predicting the entire distribution of potential, and the value of the current strongly correlates with bias. Therefore, a small network is adequate to solve this problem, eliminating over-smoothing issues that often observed in GCN-based models.

This finding reaffirms the effectiveness of our proposed approach, demonstrating that careful consideration of model complexity is necessary depending on the specificity and complexity of the prediction task at hand.

## IV. Conclusion

In summary, we have presented a universal graph-based encoding method for semiconductor devices which combines the unique characteristics of the electronic device itself and finite element meshing, revolutionizing the field of TCAD device simulation. Our innovative approach offers a unified encoding solution that can be applied to a vast range of devices, incorporating material embedding, device embedding, spatial relationship embedding, and flexible handling of self-consistent features. This comprehensive framework establishes a bridge between device simulators, such as Sentaurus TCAD, and machine learning platforms like PyTorch, unleashing the full potential of AI for EDA at the device level. Another contribution of our work is the introduction of the RelGAT model, which employs attention mechanisms to effectively capture both node and edge features, enabling a deep understanding of the intrinsic information and patterns inherent in devices. Through extensive testing and evaluation, we have demonstrated that the performance of RelGAT surpasses that of traditional GCN-based models, achieving an exceptional accuracy rate exceeding 99%. This achievement showcases the power and effectiveness of our novel model in accurately predicting device behavior.

## Data availability

The data and codes that support the results and findings within this study are available from the corresponding author upon reasonable request.

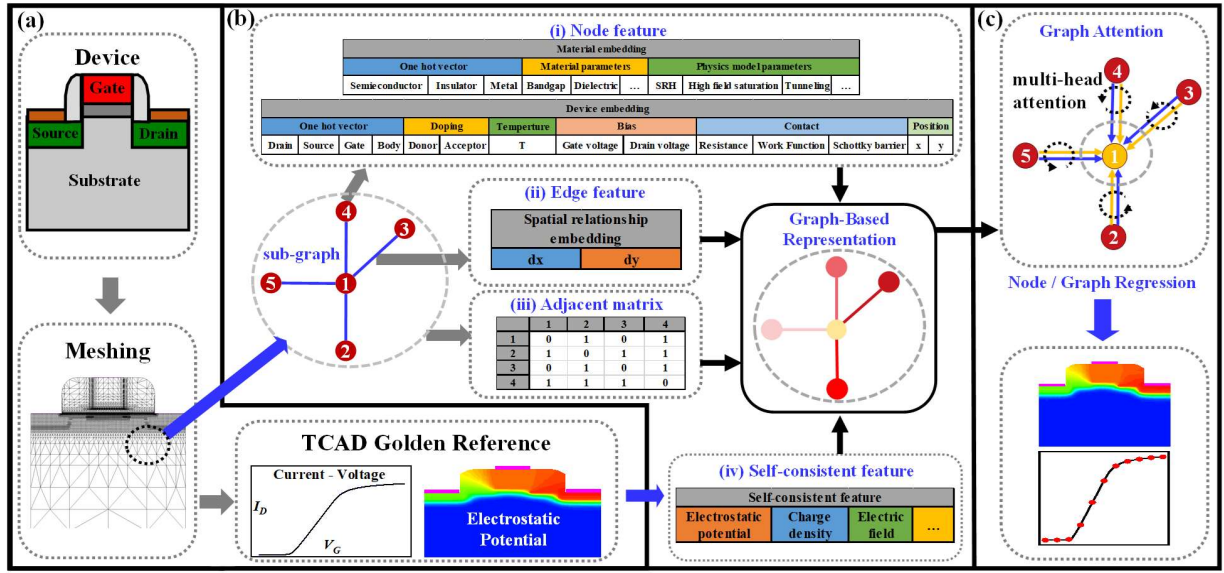

Fig. 1: Framework overview of the unified encoding scheme integrated with a graph attention network, enabling a versatile AI-driven solution for agile simulation and design of advanced devices. (a) The golden reference is obtained through TCAD simulations. (b) The comprehensive device encoding information, including material and device embedding for node feature (b.i), spatial relationship embedding for edge feature (b.ii), and adjacent matrix (b.iii), is extracted from the device meshing and TCAD material database. In addition, the self-consistent features (b.iv) are gathered from the TCAD device simulation to mimic the characteristics of the self-consistent process in TCAD simulation. (c) The graph attention-based model incorporates generic physical laws governing the device transport behavior, facilitating node-regression-based Poisson equation emulator and graph-regression-based IV predictor based on the drift-diffusion mechanism.

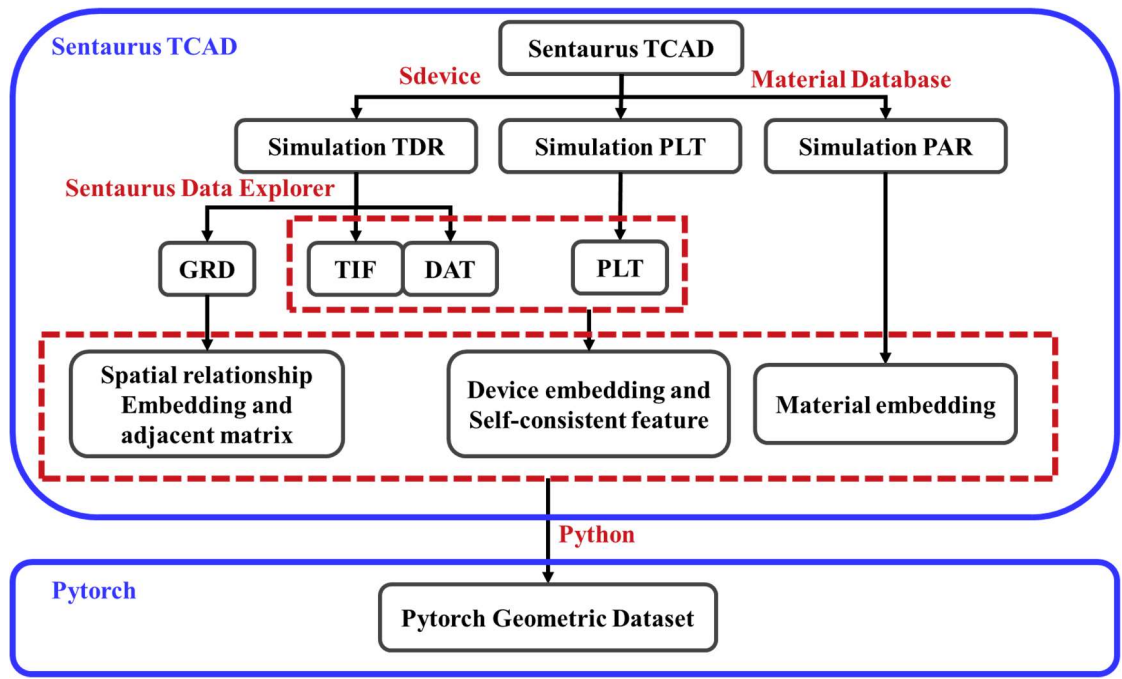

Fig. 2: Integrative API bridging device simulator (Sentaurus TCAD) and machine learning platform (Pytorch). TDR file is a proprietary format in TCAD such that it needs to be converted to a common text file format using Sentaurus Data Explorer, as they cannot be processed directly by conventional programming languages such as Python.

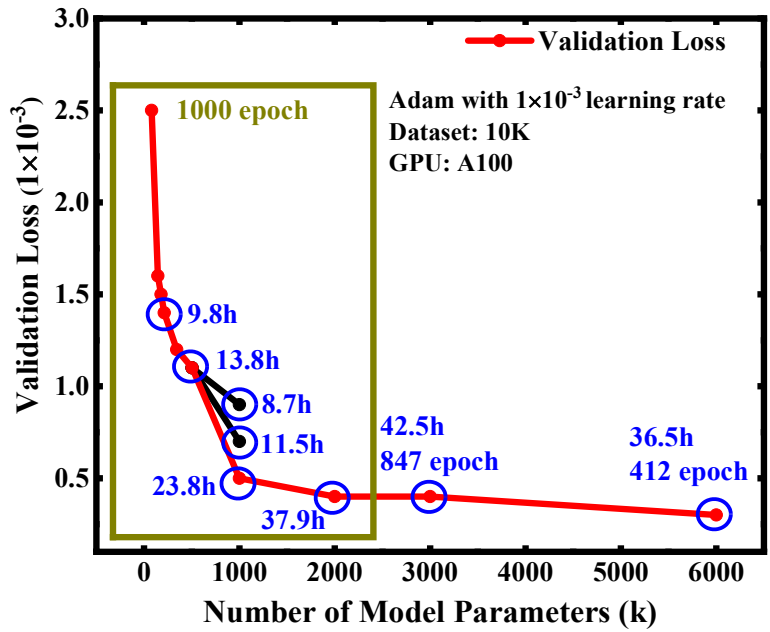

Fig. 3: Validation loss versus number of model parameters of the baseline model, illustrating a significant reduction in loss with an increasing number of model parameters, reaching saturation around 1000k. In addition, the duration of training and the number of epochs required for a model with a given number of parameters to reach a certain validation loss are also depicted. The objective is to identify a model with an optimal number of parameters, which achieves a relatively low validation error and shorter training time to serve as the baseline.

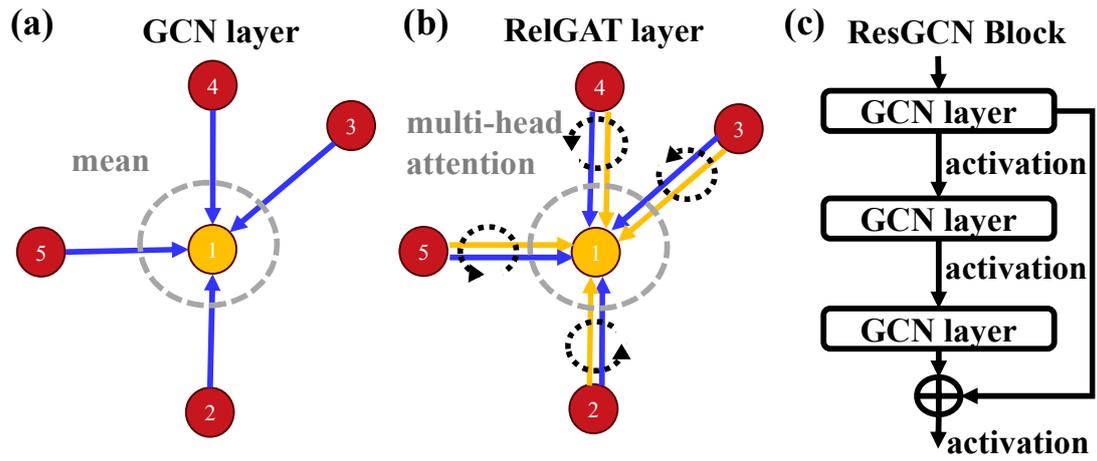

Fig. 4: Schematic illustration of the working mode of (a) GCN, (b) RelGAT, and (c) ResGCN Block.

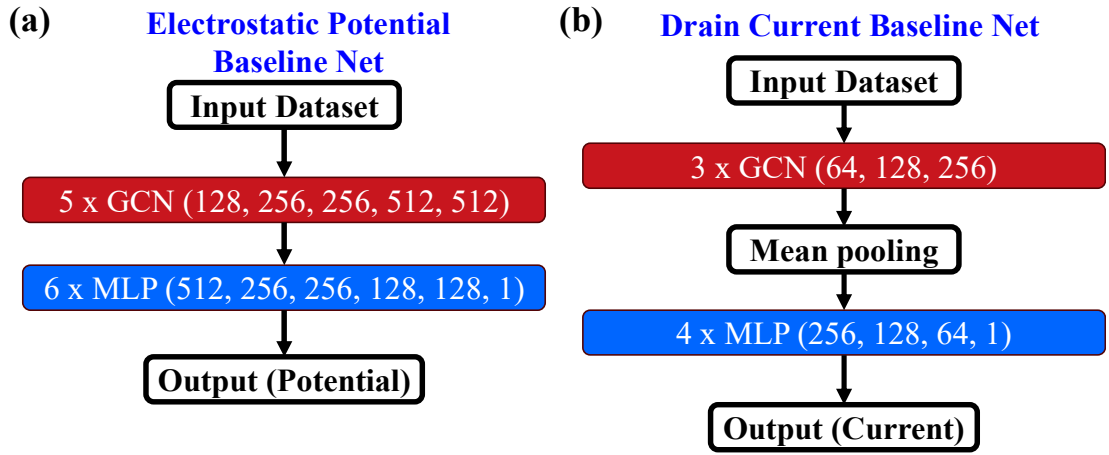

Fig. 5: The architecture of baseline models. (a) Electrostatic potential baseline net (node regression) and (b) drain current baseline net (graph regression), where each layer is followed by a layer normalization operation.

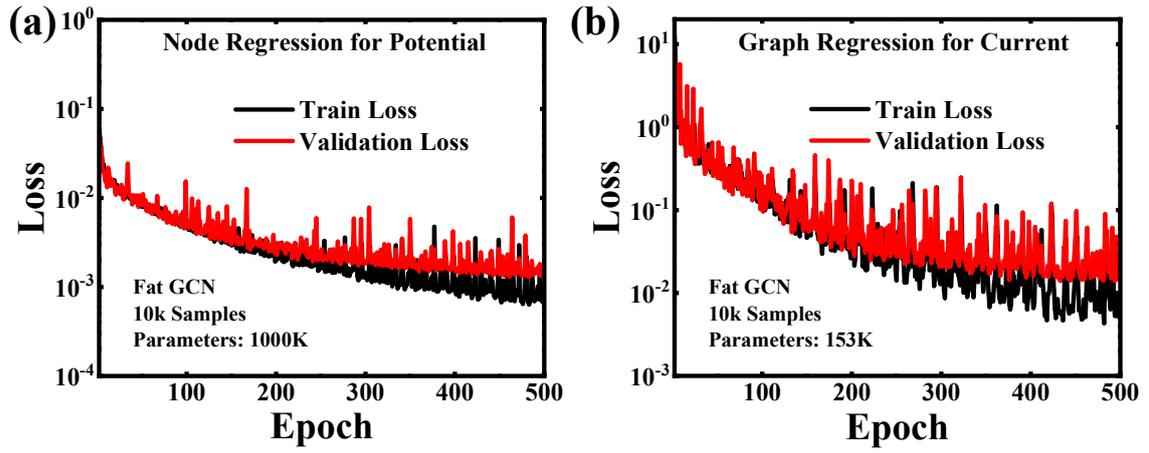

Fig. 6: Performance evaluation of baseline model during the training process with a parameter number of (a) 1000K and (b) 153k for node-regression-based Poisson emulator and graph-regression-based IV predictor, respectively.

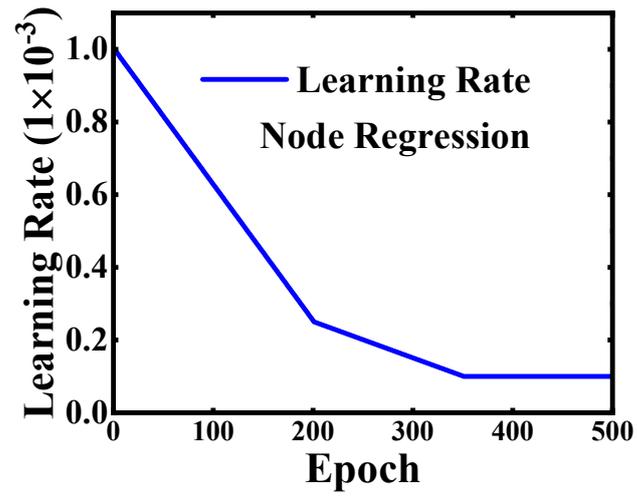

Fig. 7: Dynamic learning rate adjustment during model training for the node-regression-based Poisson emulator, facilitating rapid initial reduction of the loss function and stable fine-tuning as the optimization progresses.

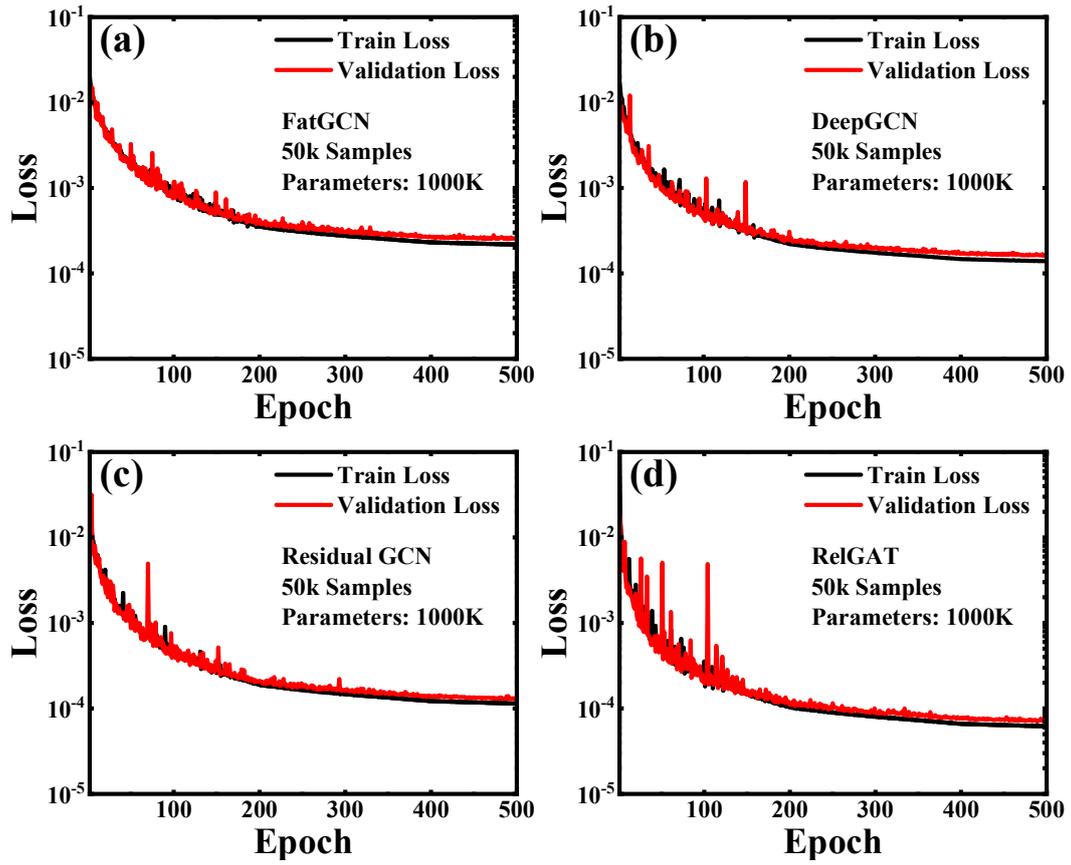

Fig. 8: Performance evaluation of different models during the training process for node-regression-based Poisson emulator: (a) FatGCN, (b) DeepGCN, (c) ResGCN, and (d) RelGAT.

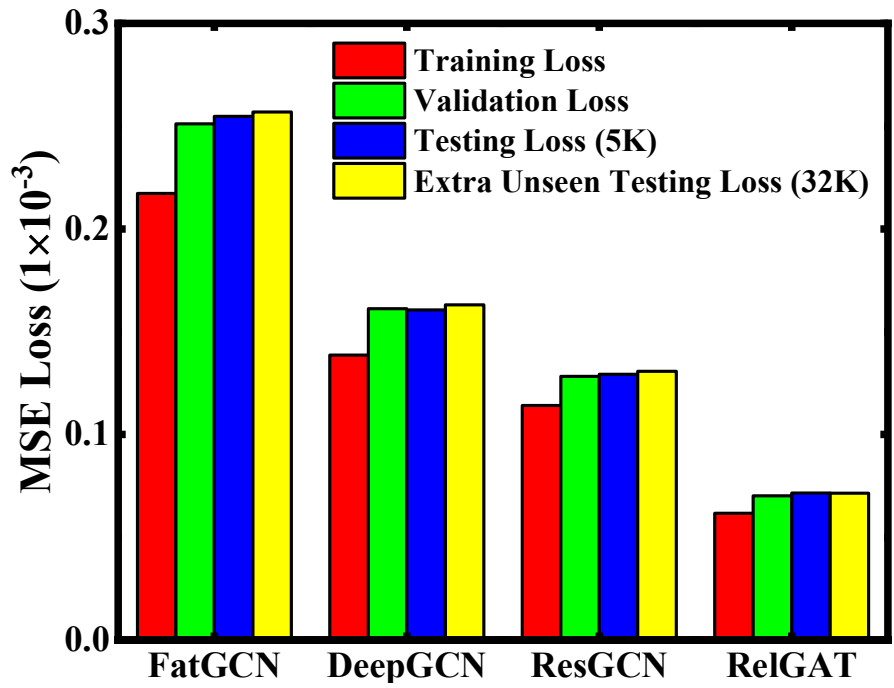

Fig. 9: Model performance based on MSE comparison during training, validation, and two unseen testing phases for all models of node-regression-based Poisson emulator.

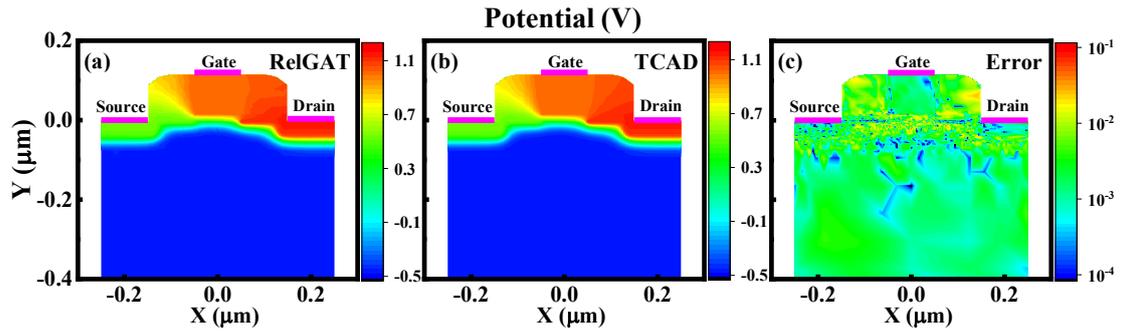

Fig. 10: Comparison of the full electrostatic potential profile obtained from (a) RelGAT, (b) TCAD, and (c) the difference between the RelGAT and TCAD predictions, demonstrating the high accuracy of our RelGAT model.

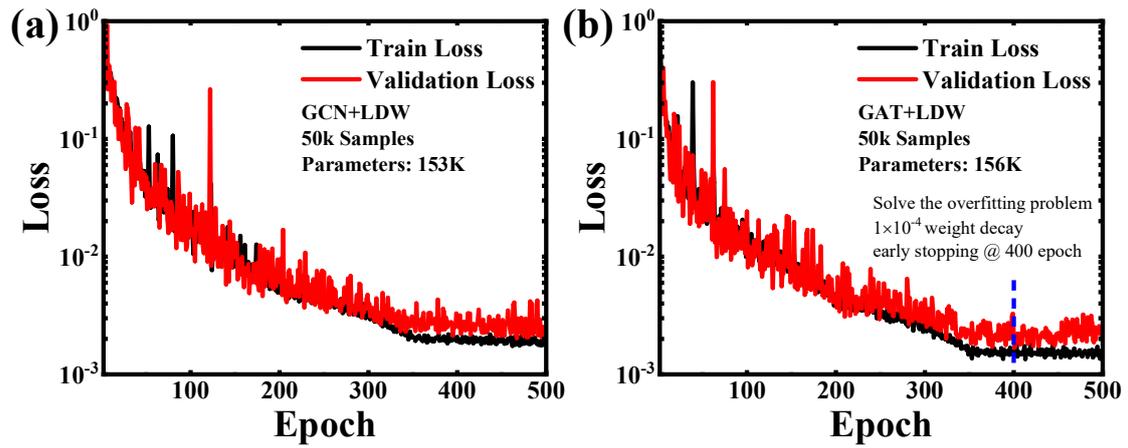

Fig. 11: Performance evaluation of different models during the training process for graph-regression-based IV predictor: (a) FatGCN, (b) RelGAT. LD means learning rate decay and W means weight decay.

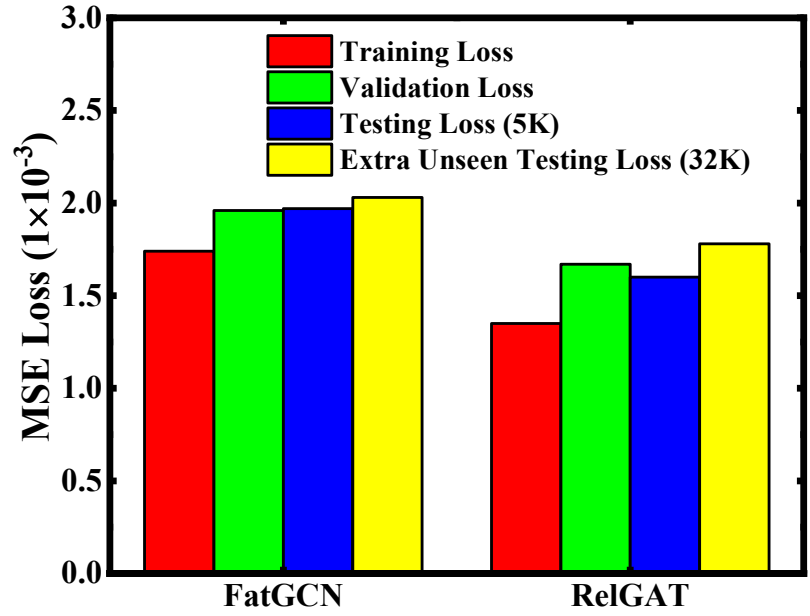

Fig. 12: Model performance based on MSE comparison during training, validation, and two unseen testing phases for FatGCN and RelGAT model of graph-regression-based IV predictor.

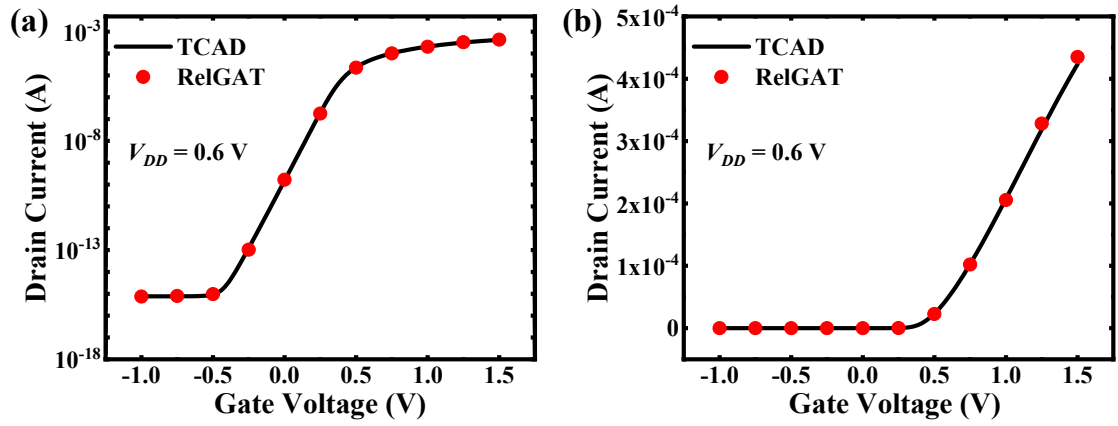

Fig. 13: Comparison of current values at different gate voltage obtained from RelGAT (red circle symbol) and TCAD (solid line) in (a) logarithm scale and (b) linear scale, showing the remarkable accuracy achieved by our RelGAT model.

Table 1: The details of architecture for different models considered in the Poisson emulator.

|  | **FatGCN** | **DeepGCN** | **ResGCN** | **RelGAT** |
|---|---|---|---|---|
| GCN/GAT Layer | 5 | 12 | 12 | 12 |
| Hidden layer (node) | 2(128),2(256),1(512) | 3(128),6(256),3(512) | 3(128),6(256),3(512) | 3(64),6(128),3(256) |
| MLP layer | 6 | 1 | 1 | 1 |
| Hidden layer (node) | 1(512),2(256),2(128),1(1) | 1(1) | 1(1) | 1(1) |
| Total layer | 11 | 13 | 13 | 13 |
| Normalization | Layernorm | Layernorm | Layernorm | Layernorm |
| Residual | None | None | Yes | None |
| Multi-head | None | None | None | 2 |
| Parameters (M) | 1.007 | 1.066 | 1.066 | 1.082 |

note: "Hidden layer (node)" notation refers to the number of hidden layers and the number of nodes in each layer. For instance, "2 (128)" means there are 2 hidden layers, and each hidden layer has 128 nodes.

Table 2: Performance Benchmarking: comparative evaluation of all models with respect to FatGCN for node-regression-based Poisson emulator on a large unseen testing set consisting of 32K random samples.

| Model | Unseen (32K) | Improvement |
|---|---|---|
| FatGCN (baseline) | $2.5691 \times 10^{-4}$ | 0% |
| DeepGCN | $1.6309 \times 10^{-4}$ | 36.52% |
| ResGCN | $1.3076 \times 10^{-4}$ | 49.10% |
| RelGAT | $7.146 \times 10^{-5}$ | 72.18% |

Table 3: The details of architecture for different models considered in the IV predictor

|  | **FatGCN** | **RelGAT** |
| --- | --- | --- |
| GCN/GAT layer | 3 | 3 |
| Hidden layer (node) | 1(64),1(128),1(256) | 1(64),1(128),1(256) |
| Pooling layer | mean | mean |
| MLP layer | 4 | 4 |
| Hidden layer (node) | 1(256),1(128),1(64),1(1) | 1(256),1(128),1(64),1(1) |
| Total layer | 7 | 7 |
| Normalization | Layernorm | Layernorm |
| Residual | None | None |
| Multi-head | None | 1 |
| Parameters (M) | 0.154 | 0.156 |

note: "Hidden layer (node)" notation refers to the number of hidden layers and the number of nodes in each layer. For instance, "2 (128)" means there are 2 hidden layers, and each hidden layer has 128 nodes.

Table 4: Performance Benchmarking: comparative evaluation of RelGAT with respect to FatGCN for graph-regression-based IV predictor on a large unseen testing set consisting of 32K random samples.

| Model | Unseen (32K) | Improvement |
|---|---|---|
| FatGCN (baseline) | 0.00203 | 0% |
| RelGAT | 0.00178 | 12.32% |